# CHISEL: Compression-Aware High-Accuracy Embedded Indoor Localization with Deep Learning

Liping Wang, Saideep Tiku, Sudeep Pasricha

*Abstract*—GPS technology has revolutionized the way we localize and navigate outdoors. However, the poor reception of GPS signals in buildings makes it unsuitable for indoor localization. WiFi fingerprinting-based indoor localization is one of the most promising ways to meet this demand. Unfortunately, most work in the domain fails to resolve challenges associated with deployability on resource-limited embedded devices. In this work, we propose a compression-aware and high-accuracy deep learning framework called *CHISEL* that outperforms the best-known works in the area while maintaining localization robustness on embedded devices.

## I. INTRODUCTION

Today's geo-location services have eliminated the need for cumbersome paper-based maps that were the dominant navigation strategy of the past. Outdoor mapping, localization, and navigation technologies have reinvented the way we interact with the world around us. However, due to the limited permeability of GPS signals within indoor environments, such services do not function in buildings such as malls, hospitals, schools, etc. In an effort to extend localization services to buildings and subterranean locales, indoor localization solutions are experiencing a recent upsurge in interest [6].

While substantial progress has been made in this area (see Section II), WiFi fingerprinting-based indoor localization stands out as the most promising solution. This is mainly due to the ubiquitous nature of WiFi Access Points (APs) and their signals in buildings, and the superior localization accuracies demonstrated with it. Fingerprinting consists of two phases: the first phase, known as the offline phase, consists of collecting WiFi signal characteristics such as RSSI (Received Signal Strength Indicator) at various indoor locations or Reference Points (RPs) in a building. The vector of wireless signal RSSI values from all APs observable at an indoor location represents a *fingerprint* of that location. Such fingerprints collected across RPs in the offline phase form a fingerprint dataset, where each row in the dataset consists of an RSSI fingerprint along with its associated RP location. Using this dataset from the offline phase, a machine learning (ML) model can then be trained and deployed on embedded devices (e.g., smartphones) equipped with WiFi transceivers. In the second phase, called the online phase, WiFi RSSI captured by a user is sent to the ML model and used to predict the user's location in the building.

Recent works report improved indoor localization accuracy through the use of Convolutional Neural Network (CNN) models [1]. This is mainly attributed to the superior ability of CNNs at discerning underlying patterns within fingerprints. Such CNN models can be deployed on smartphones and allow users to localize themselves within buildings, in real-time. Executing these models on smartphones instead of the cloud further enables security and sustainability as it eliminates the need of user data being shared through unsecured networks [2].

Unfortunately, research in the domain of indoor localization overlooks the high memory and computational requirements of CNNs, making deployment on resource-constrained embedded systems such as smartphones a challenge. While post-training model compression can ease model deployability in the online phase, it leads to an unpredictable degradation in localization performance. *Thus, there is a need for holistic deep learning solutions that can provide robust indoor localization performance when deployed on embedded devices.*

In this paper, we propose a novel multi-dimensional approach towards indoor localization that combines convolutional autoencoders, CNNs, and model compression to deliver a sustainable and lightweight framework called *CHISEL*. The main contributions of this work can be summarized as follows:

- We propose a novel RSSI pattern recognition centric, and pruning and quantization aware deep learning based indoor localization solution that combines a convolutional autoencoder (CAE) and a CNN classifier to deliver a lightweight and robust solution for embedded deployment;
- We describe a methodology for fingerprint augmentation that in combination with our proposed model improves generalization and lowers overfitting in the offline phase;
- We benchmark the performance of CHISEL against state-of-the-art ML and deep learning based indoor localization frameworks with an open indoor localization database to quantify its superior performance and lower overheads.

## II. RELATED WORK

A conventional and fairly well studied approach to indoor localization is through trilateration/triangulation using Angle of Arrival (AoA) or time of flight (ToF) methods [2]. However, AoA introduces hardware complexity and is very sensitive to computational error, especially when the distance between the transmitter and receiver becomes large. ToF needs tight synchronization requirements and even with enough resolution from signal bandwidth and sampling rate, high localization errors are common, especially when no line of sight paths are available, which is often the case in buildings. Both of these methods also require precise knowledge of AP locations, making them impractical for many indoor environments. Moreover, indoor locales pose the additional challenge of being composed of complex angular interior structures as well as diverse construction materials, e.g., concrete, metal, and wood, which contribute to hard-to-predict reductions in localization accuracy due to wireless multipath and shadowing effects [6].

The authors are with the Department of Electrical and Computer Engineering, Colorado State University, Fort Collins, Colorado 80523 USA (e-mail: liping.wang@colostate.edu; saideep@colostate.edu; sudeep@colostate.edu).

Fingerprinting based approaches have been shown to overcome many of these challenges as they do not require rigid synchronization and knowledge of AP locations, and are also less immune to multipath and shadowing effects [2]. Many RSSI fingerprinting based ML solutions have been proposed for indoor localization, e.g., approaches using Support Vector Regression (SVR) [3], k-Nearest Neighbors (KNN) [4] and Random Forest (RF) [5] based indoor location estimators.

Recent years have shown the promise of deep learning based fingerprinting methods that have outperformed classical ML approaches. Many deep learning techniques have been adapted to the domain of indoor localization [2], [6]. The work by Jang et al. [7] proposed a CNN classifier while Nowiki et al. [8] built a model consisting of a stacked autoencoder (SAE) for feature space reduction followed by a DNN classifier (SAEDNN). The experiments were conducted on the UJIIndoorLoc dataset [11] to predict the building and floor that a user is located on. However, these works do not consider positioning the user within a given floor, which is a much harder problem. At the same time, another SAEDNN based model was proposed in [9] that reduced the number of nodes in the final layer. Later, a 1-D CNN approach [10] was shown to outperform these works. This is achieved through the additional overhead of deploying separate CNN models for building, floor, and within floor prediction, which has high memory and computational costs.

*While previous works propose promising deep learning based approaches for indoor localization, they consistently overlook deployability issues arising from memory and computational overheads in embedded devices.* Post-training model compression techniques can help mitigate these deployment issues, but may lead to an unacceptable loss in localization accuracy. To overcome these challenges, we devise a novel deep learning based indoor localization framework (called CHISEL) that combines a Convolutional Autoencoder (CAE) and a CNN classifier. This model has the advantage of being amenable to model compression without notable loss in localization accuracy. This framework is discussed next.

## III. CHISEL FRAMEWORK

### A. Data Preprocessing and Augmentation

In this work, we make use of the UJIIndoorLoc indoor localization dataset [11] that covers a total of 3 buildings and 5 floors. Our approach considers a total of 905 unique RPs, such that each RP represents a unique combination of: [building ID/floor ID/space ID/relative position ID]. Here the space ID is used to differentiate between the location inside and outside a room. Relative position ID locates the user on a given floor. The RSSI values for WiFi APs vary in the range of -100 dBm to 0 dBm, where -100 dBm indicates no signal and 0 indicates full signal strength. We normalize this fingerprinting dataset to a range of 0 to 1. As there is no test data in the UJIIndoorLoc dataset, we utilize the validation component (1111 samples) of the dataset as the test set. The training portion of the dataset is split into training (15950 samples) and validation (3987 samples) subsets, based on an 80:20 split.

To compensate for the limited samples per RP and to further improve generalization, we augment the fingerprint dataset. For each RP we first calculate the mean value of all non-zero RSSI APs within one RP and the absolute difference between the mean value of each AP value. Then we generate the AP RSSI values from the uniform distribution between the difference range obtained from the first step. The final dataset is the combination of the original and augmented fingerprints.

Considering our use of convolutional deep learning networks, each fingerprint is zero-padded and translated into a single-channel square shaped image, similar to the work in [1]. For the UJIIndoorLoc dataset, this produced 24×24×1 dimensional images. This new fingerprint image-based dataset is then used to train the deep learning model described in the next subsection.

### B. Network Architecture

Table I shows our proposed deep learning model which contains the CAE and CNN components that are trained in two stages. In the first stage, a CAE comprising of an encoder and a decoder network are trained with the goal of minimizing the MSE loss between the input and the output fingerprint. This process enables the *CAE: Encoder* to efficiently extract hidden features within the input fingerprints. In the second stage, the decoder is dropped and replaced by the CNN classifier as given in Table 1. The goal of this classifier is to predict the user's location, given the encoded input from the CAE. The model is then re-trained with the weights associated with the encoder frozen in place and loss function set to sparse categorical cross-entropy. ReLU (Rectified Linear Units) is the only activation function we used for all convolutional and fully connected layers. The full model has 171209 parameters in total.

TABLE I: CHISEL's CAECNN network model layers

| Layer Type | Layer Size | Filter Count | Filter Size | Stride Value | Output Size |
|---|---|---|---|---|---|
| **CAE: Encoder** | | | | | |
| Input | — | — | — | — | 24×24×1 |
| Convolutional | 24×24 | 16 | 3×3 | 1×1 | 24×24×16 |
| Max-Pooling | — | 1 | 2×2 | 2×2 | 12×12×16 |
| Convolutional | 12×12 | 8 | 3×3 | 1×1 | 12×12×8 |
| **CAE: Decoder** | | | | | |
| Up-Sampling | — | 1 | 2×2 | 2×2 | 24×24×8 |
| Convolutional | 24×24 | 1 | 3×3 | 2×2 | 24×24×1 |
| **CNN Classifier** | | | | | |
| Convolutional | 12×12 | 8 | 3×3 | 1×1 | 12×12×8 |
| Convolutional | 12×12 | 16 | 3×3 | 1×1 | 12×12×16 |
| Max-Pooling | — | 1 | 2×2 | 2×2 | 6×6×16 |
| Convolutional | 6×6 | 32 | 3×3 | 1×1 | 6×6×32 |
| Convolutional | 6×6 | 32 | 3×3 | 1×1 | 6×6×32 |
| Max-Pooling | — | 1 | 2×2 | 2×2 | 3×3×32 |
| Flatten | 1×288 | — | — | — | 1×1×288 |
| Fully-Connected | 1×128 | — | — | — | 1×1×128 |
| Batch Norm | 1×128 | — | — | — | 1×1×128 |
| Softmax | 1×905 | — | — | — | 1×1×905 |

### C. Model Compression

We further employ a combination of two approaches to compress the parameters of our CAECNN model, for efficient deployment on embedded devices, as discussed below:

*Quantization:* The parameters of a neural network, in general, are represented by 32-bit wide floating-point values. However, a single neural network model can consist of hundreds of thousands to millions of parameters, leading to very large memory footprints and high inference latency. Quantization achieves model compression and computational relaxation by limiting the bit-width used to represent weights and activations. However, this can lead to unpredictable accuracy degradation due to a reduction in floating-point (FP) precision [12]. To overcome this issue, researchers have proposed Quantization Aware Training (QAT) which involves quantizing weights and/or activations before the training process actually begins [12]. In this work, we evaluate both post-training and training-aware quantization across a range of bit widths, and the results of this analysis are presented in section IV. We explored quantization levels ranging from 32-bits down to 2-bits and applied a uniform quantizer to all convolutional layers keeping an equal number of positive and negative representations of parameters. Scaling the input tensors can further improve quantized model accuracy [12]. We calculated scaling factors channel-by-channel using averaging absolute values of weights in each channel. In addition, to overcome the issue of vanishing gradients we applied the "straight-through estimator" (STE) [13] to the standard ReLU activation function during training.

*Pruning:* This approach involves selectively removing weights from a trained model. Out of the many pruning methodologies such as filter pruning, layer pruning, connection pruning etc., we employ connection pruning and filter pruning due to their diverse applicability and promising results across different kinds of models and layers [12]. Towards this goal, we implemented sparse connection pruning and filter pruning that are focused on zeroing-out either a weight value or entire filters based on their magnitude [14, 15]. To achieve a sparsity of *S%*, the weights of the model are ranked based on magnitude and the smallest *S%* are set to zero. In the case of filter pruning, we utilize L2-norm on the filter weights in order to rank them. We performed connection + filter pruning with varying sparsity values of 0% (no pruning), 25%, 50% and 75% for the CHISEL model to identify the best configuration, as discussed next.

## IV. EXPERIMENTS

We compare our CHISEL indoor localization framework with its data augmentation and novel CAECNN architecture with state-of-the-art deep learning based indoor localization techniques SAEDNN [9] and 1D CNN [10], as well as classical ML methods: KNN [4] and RF [5], all of which were discussed in section II. We show results for two variants of our CHISEL framework: CHISEL-DA, which uses data augmentation, and CHISEL, which does not, to highlight the impact of data augmentation. We utilized the UJIIndoorLoc dataset for all experiments. We also deployed our localization model on the Samsung Galaxy S7 smartphone to assess prediction latency.

TABLE II: Localization performance comparison

|  | KNN | RF | SAE DNN | 1D CNN | CHISEL | CHISEL -DA |
|---|---|---|---|---|---|---|
| **Building (%)** | 98.42 | 100 | 99.82 | 100 | 99.64 | 99.96 |
| **Floor (%)** | 90.05 | 91.2 | 91.27 | 94.68 | 91.72 | 93.87 |
| **Position (m)** | 9.82 | 7.85 | 9.29 | 11.78 | 8.80 | 6.95 |

### A. Evaluation on UJIIndoorLoc Dataset

The comparison of building and floor accuracy is shown in Table II. CHISEL has nearly 100% accuracy on building prediction and outperforms almost all other approaches on floor accuracy except for 1D-CNN which has three dedicated models for building, floor, and location, respectively. As shown in Table II, the best average localization (position) errors of the proposed models are ≈ 8.80 meters and ≈ 6.95 meters, respectively, for CHISEL and CHISEL-DA. Based on our analysis of the results, we believe that 1-D CNN is unable to outperform other techniques in terms of positioning accuracy due to limitations inherent in its model architecture and the lack of data augmentation, as proposed in our work.

### B. Evaluation with Model Compression

Given its better performance, we select CHISEL-DA as the baseline CHISEL model for further optimization. The uncompressed size of this CHISEL model is 801KB and delivers an average localization accuracy of 6.95 meters at a latency of 5.82 milliseconds on the Galaxy S7 device. To make the model more amenable to deployment on resource-limited embedded devices, we evaluate the impact of quantization and pruning on the accuracy, latency, and memory use of CHISEL.

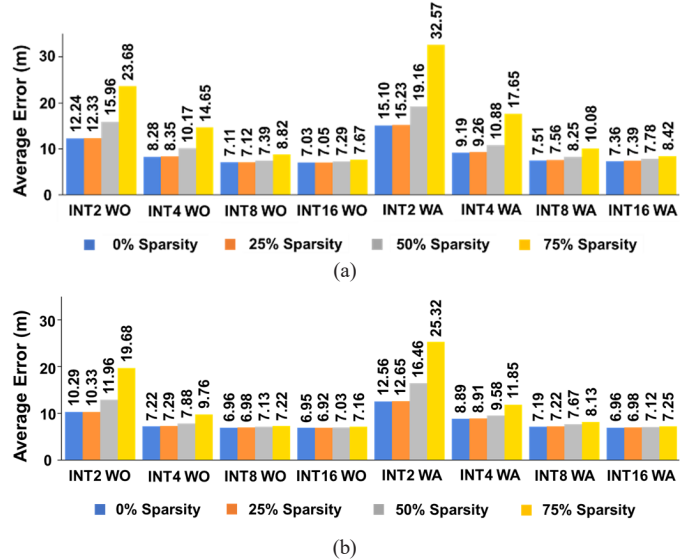

Fig. 1. Average CHISEL position error comparison across configurations with (a) Post-training quantization. (b) Quantization-aware training. Numbers on top of each bar represent the model's average error (in meters). WO and WA respectively represent weights only and weights + activations quantization.

TABLE III: Memory Footprints of Different Compression Combinations

|  | Percentage Sparsity | | | |
|---|---|---|---|---|
| **Config** | **0%** | **25%** | **50%** | **75%** |
| INT2 WO | 128KB | 116KB | 102KB | 90KB |
| INT4 WO | 173KB | 148KB | 124KB | 101KB |
| INT8 WO | 262KB | 215KB | 169KB | 122KB |
| INT16 WO | 442KB | 350KB | 259KB | 165KB |
| INT2 WA | 57KB | 46KB | 33KB | 21KB |
| INT4 WA | 107KB | 92KB | 68KB | 45KB |
| INT8 WA | 206KB | 160KB | 115KB | 68KB |
| INT16 WA | 407KB | 315KB | 222KB | 130KB |
| FP32 NQ | 801KB | 620KB | 440KB | 259K |

Fig. 1 presents the impact of various quantization and pruning configurations on the average localization error with CHISEL. The memory footprints of the CHISEL model configurations are given in Table III (FP32 NQ represents the baseline non-quantized model with 32-bit FP parameters). Configurations suffixed with WO and WA respectively represent weights-only quantization and weights+activations quantization. We summarize some of the main observations below:

1) From Fig. 1(a), we observe that post-training quantization yields models with higher localization error in all cases as compared to quantization-aware training (QAT) in Fig. 1(b). This motivates the use of QAT when deploying CHISEL.

2) As expected, an overall general trend is observed where using fewer bits to represent weight values leads to worsening accuracy. This is due to the lack of unique values available to represent weights and activations. At the extreme side, we observe that when CHISEL is about $1/17^{th}$ its original size, in the INT2-WA-25% configuration (46KB) (Fig. 1(b)) it makes localization error $\approx 1.82\times$ larger than CHISEL-DA but is still competitive with 1-D CNN (Table II).

3) Pruning from 0% (no pruning) to 25% has almost no impact on localization accuracy while reducing the model footprint by up to 25% as seen for the INT16-WA configurations in both Fig. 1(a) and 1(b). This is strongly suggestive of pruning's positive impact towards deep-learning model deployment for indoor localization with CHISEL.

4) The impact of pruning becomes more pronounced when aggressively quantizing the model. This is especially true for the WA quantization as shown in Fig. 1(b). It is important to pay more attention to activations when quantizing CNN models with low bits, or aggressive quantization may result in huge accuracy reduction after compression.

5) Based on the results observed in Fig. 1 and Table III, a good candidate for the compressed model is INT4-WO-25% with QAT, resulting in a 148KB memory footprint. This model is $\approx 5.41\times$ smaller than baseline CHISEL model and still better in terms of accuracy than classical ML models [4, 5] as well as state-of-the-art deep learning models in prior works [9, 10].

Lastly, to capture the impact of compression on localization latency, we deployed all of the compressed configurations of CHISEL on a Samsung Galaxy S7 smartphone using Android Studio's on-device inference framework in v4.0.1. Our application is designed to directly receive RSSI from a file that contains all 1111 samples from the test set. RSSI values are processed into matrices in-app and fed to the CHISEL model. The captured latencies include the time required to pre-process the RSSI fingerprint into images and are averaged over 100 repetitions.

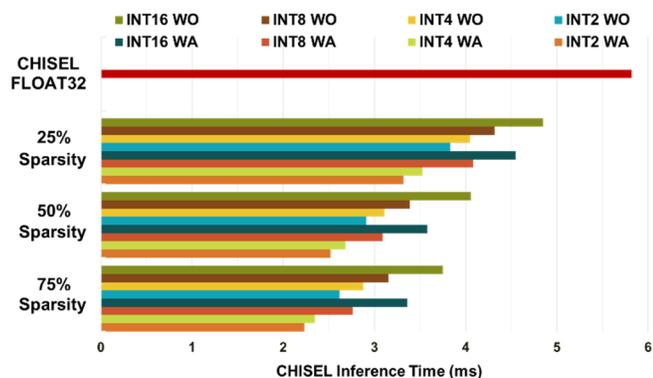

Fig. 2. On-device inference time from all compression configurations of CHISEL. CHISEL FLOAT32 represents the baseline model.

Inference time results of all compression configurations are presented in Fig. 2. We observe that both quantization and pruning can offer notable acceleration over the baseline FLOAT32 models. The INT4-WA-50% sparsity model cuts localization latency to half (~2.8ms), while taking a penalty of 2.63m (38%) in terms of positioning error. Aggressive quantization and pruning beyond this point yields limited benefits, e.g., INT2-WA+75% sparsity only reduces the latency to ~2.25ms while degrading the localization accuracy by $3\times$. INT4-WO-25% continues to present itself as a good candidate with a notable ~31% reduction in latency.

*In summary,* the intelligent data-augmentation and novel CAECNN deep learning network model which is amenable to model compression allows our CHISEL framework to provide new options for high accuracy indoor localization while minimizing deployment costs (in terms of memory footprint and latency) on embedded devices.

## V. CONCLUSION

In this paper, we presented a novel indoor localization framework called CHISEL. Our approach outperforms state-of-the-art ML and deep learning-based localization frameworks, achieving higher localization accuracy. The compression-friendly CAECNN models in CHISEL can maintain robust accuracy in the presence of aggressive model compression. Our compressed model versions are easy to deploy on smartphones and resource-constrained embedded devices that may have KB-sized resource budgets. Based on our experimental analysis, CHISEL is shown to provide a spectrum of deployment configurations with varying tradeoffs between accuracy, memory footprint, and latency goals. One of the more promising CHISEL configurations is the INT4-WO-25% with QAT, which reduces model size to 148KB (81.52% reduction) and reduces latency by 1.80ms (30.93% reduction) at the cost of sacrificing 0.34m (4.89%) localization accuracy.